\documentclass[conference]{IEEEtran}
\IEEEoverridecommandlockouts

\usepackage[utf8]{inputenc} 
\usepackage[T1]{fontenc}    
\usepackage{hyperref}       
\usepackage{url}            
\usepackage{booktabs}       
\usepackage{amsfonts}       
\usepackage{nicefrac}       
\usepackage{microtype}      
\usepackage{xcolor}         

\usepackage{multirow}
\usepackage{todonotes}
\usepackage{amsmath}
\usepackage{amssymb}
\usepackage{graphicx}
\usepackage{caption}
\usepackage{subcaption}
\usepackage{array}
\usepackage{makecell}
\usepackage{adjustbox}
\usepackage{algorithm}
\usepackage{algpseudocode}
\usepackage{tikz}
\usepackage{placeins}
\usepackage{pgfplots}
\usepackage{pgfplotstable}
\pgfplotsset{compat=1.18}
\usepgfplotslibrary{colormaps}
\usepgfplotslibrary{patchplots}
\usepgfplotslibrary{groupplots}
\usepgfplotslibrary{statistics} 
\usetikzlibrary{calc}
\usepackage{multirow}
\usepackage{makecell}  

\begin{document}

\title{CRISP: A Framework for Clause-Reconstructed Interpretable NeuroSymbolic Propositions}

\author{%
    \IEEEauthorblockN{
        Alex Chan\IEEEauthorrefmark{1}, 
        Shafi Muhtasim Chowdhury\IEEEauthorrefmark{2}, 
        Ekin Can Erkus\IEEEauthorrefmark{1}, 
        Ole-Christoffer Granmo\IEEEauthorrefmark{3}, \\
        Alex Yakovlev\IEEEauthorrefmark{1}, 
        and Rishad Shafik\IEEEauthorrefmark{1}
    }
    \IEEEauthorblockA{\IEEEauthorrefmark{1}School of Engineering, Newcastle University, Newcastle upon Tyne, UK}
    \IEEEauthorblockA{\IEEEauthorrefmark{2}Northumbria University, Newcastle upon Tyne, UK}
    \IEEEauthorblockA{\IEEEauthorrefmark{3}Centre for Artificial Intelligence Research, University of Agder, Grimstad, Norway}
    \small{\{alex.chan, ekin.erkus, alex.yakovlev, rishad.shafik\}@newcastle.ac.uk, shafi.chowdhury@northumbria.ac.uk, ole.granmo@uia.no}
    \thanks{\textbf{Arxiv preprint version. This is an accepted paper for the International Symposium on the Tsetlin Machine (ISTM) 2026.}}
}

\maketitle
\begin{abstract}
Deep neural networks achieve high accuracy through layered numerical transformations, yet their decisions remain difficult to audit because decision evidence is encoded in hidden activations rather than explicit rules. This paper introduces CRISP, a framework that reconstructs the last-layer activation vector (LLAV) of binary neural teachers as Tsetlin Machine (TM) clauses. CRISP sign-binarizes the teacher's penultimate pre-logit activations, and assigns one Individual TM (ITM) to each LLAV neuron. Each reconstructed hidden bit is represented by propositional clauses over Booleanized input features, which gives a direct symbolic trace from named input thresholds to a named teacher neuron. CRISP is evaluated on MNIST, KMNIST, FashionMNIST (FMNIST), SVHN, and CIFAR10 using a BinaryConnect convolutional neural network (BCCNN) teacher and a fully binary neural network (BNN) teacher, with an additional study on binary thresholding, thermometer encoding, and quartile binning at multiple bit depths. The results show that LLAV sign-binarization does not reduce teacher-head accuracy in the tested BNN setting, while ITM reconstruction error is the main limiting factor. Quartile one-bit Booleanization gives the strongest reconstruction fidelity on SVHN at 87.52\% test fidelity and is competitive on CIFAR10, and the reconstructed LLAV preserves 78.41\% teacher-head accuracy on FMNIST. Pooled clause-evidence visualizations show that the learned ITM literals concentrate on the object region in centered benchmarks. CRISP therefore provides a clause-level route for inspecting the final hidden representation of binary neural teachers.
\end{abstract}


\section{Introduction}\label{sect:introduction}

Deep neural networks~\cite{lecun2015deep} are widely used for image classification, medical decision support, financial risk modeling, and other high-dimensional tasks~\cite{schmidhuber2015deep, gu2020empirical}. Their accuracy comes from hierarchical representation learning, where early layers extract local evidence, deeper layers compose this evidence into task-relevant abstractions, and final hidden states concentrate decision information into compact internal codes. However, this representational depth also creates an auditing problem, as the final prediction is not completely visible. This is because the path to that prediction is not human-understandable, as it is distributed across many hidden units and parameters that are expressed through arithmetic transformations. As a result, the learned model is difficult to inspect, verify, or translate into symbolic form~\cite{lipton2018mythos,tarasyuk2025learning,erkucs2024artificial}.

Structurally, a neural network is a composition of matrix multiplications, additions, normalizations, and nonlinearities~\cite{lecun2015deep}. The final prediction emerges from the cumulative effect of these operations, while many intermediate states are difficult to assign direct semantic meaning~\cite{lipton2018mythos,rudin2019stop,montavon2018methods}. The pre-logit activations in the final hidden layer are a natural target for symbolic reconstruction because they summarize the computation immediately before classification~\cite{yosinski2014transferable,alain2016understanding}. Prior work shows that deeper layers capture more class-discriminative features~\cite{bengio2013representation}, final-layer visualizations reveal structured semantic transformations~\cite{zeiler2014visualizing}, and final hidden representations often contain linearly separable class information~\cite{alain2016understanding}. Decoding this layer in terms of the original input therefore offers a focused way to study what the network has learned without reconstructing every intermediate operation.

This paper proposes CRISP, a framework that pairs a binary neural teacher with a bank of Tsetlin Machines (TMs)~\cite{2018-arxiv-granmo-tsetlin}. Each Individual Tsetlin Machine (ITM) reconstructs one coordinate of the teacher's binary last-layer activation vector (LLAV) from a Boolean encoding of the original input. Each reconstructed LLAV bit is expressed by propositional clauses over binarized input features. The resulting surrogate preserves neuron identity because every hidden coordinate remains tied to its assigned ITM, and it provides an inspection path because each ITM exposes the literals that support or oppose the reconstructed firing state.

CRISP tests the hypothesis that the internal firing pattern of a black-box binary neural network can be approximated by symbolic clauses over the original input. This hypothesis leads to three research questions. \textit{RQ1}) Can symbolic LLAV surrogates reconstruct a teacher network's pre-logit activations despite the cumulative effect of successive filtering and transformations? \textit{RQ2}) How does the input Booleanization strategy affect reconstruction quality? \textit{RQ3}) How does teacher architecture affect symbolic reconstruction fidelity?

To answer these questions, CRISP is evaluated with two binary teacher architectures, a fully binarized neural network (BNN)~\cite{2016-arxiv-courbariaux_et_al-binary_neural_networks,hubara2016bnn,qin2020binary} and a BinaryConnect Convolutional Neural Network (BCCNN)~\cite{courbariaux2015binaryconnect}. The main experimental flow spans five image classification datasets, namely, MNIST~\cite{1998-IEEE-lecun_et_al-MNIST}, KMNIST~\cite{clanuwat2018kmnist}, FashionMNIST (FMNIST)~\cite{xiao2017fmnist}, SVHN~\cite{2011-NIPS-netzer_et_al-SVHN}, and CIFAR10~\cite{2009-tech_report-krizhevsky-CIFAR10}, to provide broader validity of the results. A Booleanization study then compares binary thresholding, thermometer encoding, and quartile binning at multiple bit depths on SVHN and CIFAR10.

\begin{figure*}[t]
    \centering
    \includegraphics[width=0.85\linewidth]{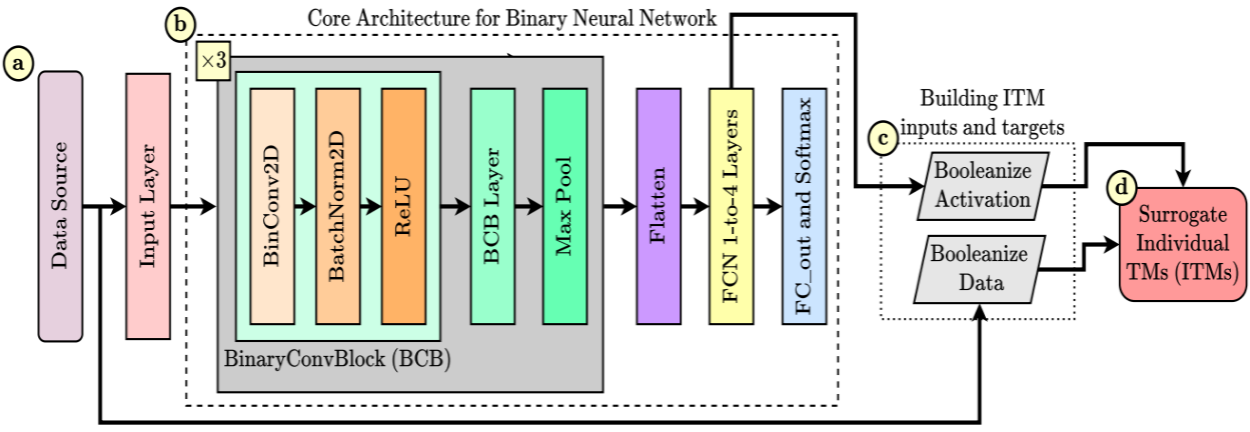}
    \caption{CRISP workflow. A binary neural teacher is trained on the image task, the teacher's LLAV is extracted and sign-binarized, the original input is Booleanized, and one ITM is trained per LLAV coordinate to form a clause-based hidden-state surrogate.}
    \label{fig:crisp_pipeline}
    \vspace{-7.5px}
\end{figure*}

Thus, this paper makes three contributions. \textit{First}, it defines a neuron-wise symbolic surrogate layer that aligns one ITM with one teacher hidden neuron and reconstructs the LLAV from Booleanized input. \textit{Second}, it quantifies how input Booleanization and teacher architecture affect symbolic reconstruction fidelity. \textit{Third}, it shows how reconstructed LLAV bits can be inspected as clauses over named pixel-threshold features, including pooled clause-evidence visualizations that link hidden-bit reconstruction to input regions.
CRISP is intended as a representation-level audit tool rather than a classifier: it exposes the teacher's final hidden interface as inspectable clauses rather than replacing the teacher head, and targets the LLAV since that is the representation the binary classifier head consumes.

Such an LLAV reconstruction offers two benefits that are not usually available together in per-prediction explanation tools. 
First, the ITMs are fixed after training, so the same clauses can be inspected across samples, classes, and model variants. The surrogate is therefore a persistent audit object rather than a single local explanation. 
Second, because the reconstruction targets the exact binary interface the classifier head consumes, the reconstructed binary activations can be fed directly through the teacher's own head to test whether it retains class-relevant information, without rerunning the full teacher. 
CRISP therefore evaluates both interpretability and representational sufficiency since clauses expose how hidden bits are reconstructed, while the injection test measures whether the reconstructed hidden states remain usable for classification.

Section~\ref{sect:crisp} defines CRISP and positions it relative to prior work. Section~\ref{sect:methods} describes the experimental protocol. Section~\ref{sect:results} presents baseline reconstruction results, the LLAV injection test, the Booleanization study, and clause-level interpretation examples. Section~\ref{sect:conclusion} concludes the paper.

\section{CRISP: Proposed Framework}
\label{sect:crisp}
Figure~\ref{fig:crisp_pipeline} depicts the CRISP workflow, which operates in two stages. The first stage trains a binary neural teacher on the classification task and extracts the binarized LLAV for each sample. The second stage trains one ITM per LLAV neuron, where each ITM learns to reconstruct its assigned neuron's firing state from a Boolean encoding of the original input. The output is a symbolic surrogate of the teacher's final hidden representation in which each reconstructed bit corresponds to inspectable clauses over input features.

The teacher first maps the input to a pre-logit representation. This representation is converted into the LLAV and paired with a Booleanized version of the original input to build the training set for the ITM bank. 
Representational sufficiency is assessed by an injection test in Section~\ref{sect:results:llav_injection}, where the reconstructed LLAV is passed through the teacher's own classification head $fc_out$ to measure whether the reconstructed neurons are usable for classification without introducing a separately trained classifier.
The framework is therefore centered on symbolic hidden-state reconstruction. Input Booleanization defines the clause vocabulary, LLAV extraction defines the reconstruction targets, and the ITM bank defines the symbolic surrogate.

Algorithm~\ref{alg:crisp} summarizes the training and inference procedure. The teacher is trained first and then frozen. LLAV targets are extracted once for the training, validation, and test splits. Each ITM is trained independently against one LLAV coordinate, so the 256-neuron setting produces 256 binary symbolic reconstruction tasks. At inference time, a new input is Booleanized, passed through the ITM bank, and mapped to a reconstructed LLAV that can be inspected directly.

\begin{algorithm}[h]
    \caption{CRISP Symbolic LLAV Reconstruction}
    \footnotesize
    \label{alg:crisp}
    \begin{algorithmic}[1]
    \Require Training data $\mathcal{D}_{\mathrm{tr}}$, validation data $\mathcal{D}_{\mathrm{val}}$, test data $\mathcal{D}_{\mathrm{te}}$, teacher $f_T$, Booleanizer $\phi$, hidden size $H$
    \Ensure ITM bank $\{M_j\}_{j=1}^{H}$ and reconstructed LLAV $\hat{h}(x)$
    \State Train binary teacher $f_T$ on $\mathcal{D}_{\mathrm{tr}}$ and select hyperparameters using $\mathcal{D}_{\mathrm{val}}$.
    \State Freeze $f_T$ and extract $h_T(x)$ from Equation~\eqref{eq:hidden_binary} for all splits.
    \State Fit any data-dependent Booleanization parameters on $\mathcal{D}_{\mathrm{tr}}$ only.
    \State Transform each input to $\phi(x)$ for training, validation, and test samples.
    \For{$j=1$ to $H$}
        \State Train ITM $M_j$ to predict $h_{T,j}(x)$ from $\phi(x)$.
    \EndFor
    \State Form $\hat{h}(x)$ using Equation~\eqref{eq:itm_bank}.
    \State Evaluate reconstruction fidelity with Equation~\eqref{eq:reconstruction}.
    \end{algorithmic}
\end{algorithm}

\subsection{Relation to Prior Work}
\label{sect:crisp:related}

Interpretability methods for neural networks can be grouped by what they explain. Attribution and saliency methods assign importance to input features for a given prediction~\cite{simonyan2014saliency,lundberg2017shap}. Local surrogate methods approximate a model around a sample or a local neighborhood~\cite{ribeiro2016lime}. Concept-based approaches connect predictions to user-defined concepts~\cite{koh2020concept}. Visualization methods inspect filters, activations, or internal representations~\cite{zeiler2014visualizing}. These approaches can identify relevant associations, but they do not convert a named hidden coordinate into a persistent symbolic object, which can be reused across samples and directly tested as an input to the classifier head.

TMs provide a different form of interpretability because their clauses are explicit conjunctions over Boolean literals~\cite{2018-arxiv-granmo-tsetlin}. This property makes TMs well suited for rule-like symbolic representations, but standard TM classifiers operate directly on labels rather than on internal neural states. Related symbolic-transfer work has approximated final hidden activations with symbolic regression~\cite{shmuel2026interpretable}, and TM-based distillation has used teacher output distributions to guide TM students~\cite{kinateder2026distribution}. 
CRISP differs in target and granularity. It reconstructs the BNN's pre-logit LLAV neuron by neuron, and each reconstructed bit is represented as a set of TM clauses over named input literals. 
In particular, unlike temperature-scaled logit distillation, CRISP reconstructs the hard, sign-binarized LLAV instead of a softened output distribution, so there is no softmax temperature to tune.
The explanation is therefore tied to the teacher's internal representation rather than only to the output label. This representation-level use of TM clauses also connects CRISP to the wider neural-symbolic integration literature, which argues for learning systems whose internal reasoning remains legible as models scale~\cite{garcez2023neurosymbolic}.

\subsection{Problem Setting and Teacher Extraction}
\label{sect:crisp_problem}

Let $x\in\mathcal{X}$ denote an input image and $y\in\{1,\ldots,C\}$ denote its ground-truth class. A teacher $f_T$ produces a predicted label $\tilde{y}(x)=\arg\max_k p_{T,k}(x)$ and a binary hidden state $h_T(x)\in\{0,1\}^{H}$ at the penultimate layer. The binary LLAV can be obtained as
\begin{equation}
h_T(x)=\mathbf{1}[g_T(x)>0],
\label{eq:hidden_binary}
\end{equation}
where $g_T(x)\in\mathbb{R}^{H}$ is the pre-threshold activation vector of the teacher's penultimate layer, and the indicator is applied coordinate-wise. As shown in Figure~\ref{fig:crisp_pipeline}b--c, $h_T(x)$ is extracted after the teacher has combined convolutional filtering, pooling, and fully connected binary transformations, so each LLAV coordinate can aggregate spatially extended evidence from the input~\cite{zeiler2014visualizing}. In the reported experiments, $H=256$.

CRISP constructs a symbolic surrogate $\hat{h}(x)$ that reproduces $h_T(x)$ while exposing a clause-level trace from each reconstructed bit to the input. Reconstruction quality for neuron $j$ is measured by
\begin{equation}
q_j= \frac{1}{N}\sum_{i=1}^{N}\mathbf{1}\!\left[\hat{h}_j(x_i)=h_{T,j}(x_i)\right],
\label{eq:reconstruction}
\end{equation}
where $N$ is the number of evaluation samples, $h_{T,j}(x_i)$ is the teacher's binary activation for neuron $j$, and $\hat{h}_j(x_i)$ is the corresponding ITM reconstruction. The aggregate reconstruction fidelity is the mean of $q_j$ over all $H$ neurons, and the reconstruction fidelity measures how well the symbolic bank reproduces the neural teacher's binary activations, with an injection test in Section~\ref{sect:results:llav_injection} that separately determines if the reconstructed activations are usable for classification. 

The LLAV binarization in Equation~\eqref{eq:hidden_binary} can discard activation magnitude. In the BNN injection experiment reported in Section~\ref{sect:results:llav_injection}, this loss does not reduce teacher-head accuracy because the final binary output layer applies a sign operation before classification. Thus, the tested BNN head already uses the LLAV sign pattern as its operative interface. CRISP takes this binary interface as the reconstruction target and asks whether it can be recovered by clauses over the original input.

\subsection{Input Booleanization}
\label{sect:crisp:booleanization}

ITM training requires each input to be expressed as a Boolean vector because TM clauses are defined over literals and their negations~\cite{2018-arxiv-granmo-tsetlin}. Let $\phi(x)\in\{0,1\}^{d}$ denote the Booleanized input, where $d$ is the encoded input dimension. The encoding scheme determines both the information retained from the raw input and the clauses that the ITM can form. CRISP evaluates three schemes at bit depths $n_{\text{bits}}\in\{1,2,4\}$ in the Booleanization study: Binary thresholding, thermometer encoding, and quartile binning.

For a normalized scalar feature $u\in[0,1]$, binary thresholding produces one bit, which serves as a one-threshold baseline:
\begin{equation}
b(u) = \mathbf{1}[u > 0.5].
\label{eq:binary}
\end{equation}

Thermometer encoding first maps $u$ to an integer level,
\begin{equation}
\ell(u)=\min\!\left(\left\lfloor u\, n_{\text{bits}}\right\rfloor,\ n_{\text{bits}}-1\right),
\label{eq:therm_level}
\end{equation}
where $\ell(u)$ is the quantized level. 
The encoded bits are then
\vspace{-2.5px}
\begin{equation}
    t_m(u)=
    \begin{cases}
        \mathbf{1}[u>0.5], & n_{\text{bits}}=1,\\
        \mathbf{1}[\ell(u)>m], & n_{\text{bits}}\geq2,
    \end{cases}
    \label{eq:therm}
\end{equation}
for $m=0,\ldots,n_{\text{bits}}-1$, where $t_m(u)$ is the $m$-th thermometer bit.
This representation preserves ordinal threshold information, so a clause can express that a feature exceeds a level using a single literal~\cite{buckman2018thermometer}.
Note that the top thermometer bit $m = n_{bits}-1$ evaluates $\mathbf{1}[\ell(u) > n_{bits}-1]$, which is never satisfied due to $\ell(u) \le n_{bits}-1$. Therefore, it is constant (always $0$) and non-informative, so an $n_{bits}$-bit thermometer code carries $n_{bits}-1$ informative bits.

Quartile binning replaces the uniform levels in Equation~\eqref{eq:therm_level} with data-dependent quantile thresholds estimated from the training set only. Let $q_1,\ldots,q_{n_{\text{bits}}-1}$ denote the $\frac{1}{n_{\text{bits}}},\ldots,\frac{n_{\text{bits}}-1}{n_{\text{bits}}}$ empirical quantiles of a feature. The bin index is
\vspace{-2.5px}
\begin{equation}
    k(u)=\sum_{r=1}^{n_{\text{bits}}-1}\mathbf{1}[u>q_r],
\label{eq:quartile_bin}
\end{equation}
The final quartile bits are
\begin{equation}
q_m(u)=
\begin{cases}
\mathbf{1}[u>q_1], & n_{\text{bits}}=1,\\
\mathbf{1}[k(u)>m], & n_{\text{bits}}\geq2,
\end{cases}
\label{eq:quartile_enc}
\end{equation}
for $m=0,\ldots,n_{\text{bits}}-1$, where $q_m(u)$ is the $m$-th quartile bit and, for $n_{\text{bits}}=1$, no interior quantiles are defined, so $q_1$ is taken as
the training-set median of the feature to be consistent with the one-bit setting used in Section~\ref{sect:Booleanization}.
Quartile binning adapts thresholds to the empirical feature distribution instead of assuming that intensities are uniformly spread over $[0,1]$~\cite{gronningsater2024toolbox}. All data-dependent thresholds are fitted on training data only and then applied unchanged to validation and test data.

\subsection{ITM Bank Training and Inference}
\label{sect:crisp:itm}

CRISP trains one ITM $M_j$ for each hidden neuron $j\in\{1,\ldots,H\}$, as illustrated in Figure~\ref{fig:crisp_pipeline}c--d. Each ITM predicts whether its assigned teacher neuron fires:
\begin{equation}
\hat{h}_j(x)=M_j\!\left(\phi(x)\right),
\label{eq:itm}
\end{equation}
where $\phi(x)$ is the Booleanized input. The surrogate bank then produces
\begin{equation}
\hat{h}(x)=\left[\hat{h}_1(x),\ldots,\hat{h}_H(x)\right].
\label{eq:itm_bank}
\end{equation}
Each clause in $M_j$ is a conjunction over literals from $\phi(x)$ and their negations. A reconstructed bit $\hat{h}_j(x)=1$ therefore corresponds to a set of active clauses, and each active clause identifies a specific input-threshold pattern associated with the teacher neuron's activation.

\subsection{Interpretability Scope and Computational Cost}
\label{sect:crisp:cost}

The symbolic part of CRISP is the ITM bank, where the teacher remains as a learned neural model. CRISP should therefore be interpreted as a representation-level reconstruction method rather than a claim of full teacher transparency.

The reconstruction metric in Equation~\eqref{eq:reconstruction} measures whether the ITM bank reconstructs the teacher LLAV. The clause representation also allows each reconstructed bit to be inspected, which is possible because Booleanization assigns a known meaning to every input bit, as each literal relates to a feature-threshold statement instead of an anonymous array index. Thus, CRISP is neuro-symbolic in the sense that the reconstruction target originates from a neural teacher, and the reconstructed form is a set of symbolic clauses over named features~\cite{bhuyan2024neuro}.

For a given sample, clause evidence is pooled across all $H$ ITMs and mapped back to the input coordinates constrained by the surviving literals. Each literal is classified along two axes, namely, clause polarity, which indicates whether the clause supports or opposes its ITM output, and literal polarity, which indicates whether the literal demands a thresholded feature $X_i$ or its negation $\overline{X_i}$. Under thermometer encoding, the top threshold bit is always false by construction in Equation~\eqref{eq:therm}. Its negation is therefore excluded from the pooled maps so that a fixed, non-informative term does not dilute the evidence. Section~\ref{sec:explainability} applies this procedure to representative samples.

Let $H$ denote the number of LLAV coordinates, $E_{\mathrm{ITM}}$ the number of ITM training epochs, $m$ the number of clauses per ITM, and $d$ the Booleanized input dimension. A conservative worst-case training-pass cost scales as
\begin{equation}
\mathcal{O}\!\left(H E_{\mathrm{ITM}} m d\right),
\label{eq:crisp_cost}
\end{equation}
where the dominant term comes from evaluating clause literals across the ITM bank. This cost is parallel over LLAV coordinates, as each ITM is trained independently. The method introduces an offline reconstruction cost, but keeps inference and inspection modular, so that each reconstructed bit can be evaluated, inspected, or replaced without retraining the bank.

\section{Experimental Setup}
\label{sect:methods}

\subsection{Data}
\label{sect:methods:datasets}

Five ten-class image classification benchmarks are used as described in Table~\ref{tab:datasets}, which summarizes the image size, channel count, and the training, validation, and test split of each dataset. The benchmark set spans grayscale digit recognition, visually similar handwritten character recognition, clothing classification, street-view digit recognition, and color object classification. This range tests whether LLAV reconstruction remains stable as the teacher representation becomes more spatially and semantically demanding. Binary network architectures are used as teachers because their classifier heads expose sign-based hidden interfaces, so the LLAV captures the state that the binary classifier consumes.

\begin{table}[h]
\centering
\caption{Summary of the five benchmark datasets used.}
\label{tab:datasets}
\footnotesize
\begin{tabular}{lccccc}
\toprule
Dataset & Size & Channels & Train & Val & Test \\
\midrule
MNIST     & $28\times28$ & 1 & 55,000 & 5,000 & 10,000 \\
KMNIST    & $28\times28$ & 1 & 55,000 & 5,000 & 10,000 \\
FMNIST    & $28\times28$ & 1 & 55,000 & 5,000 & 10,000 \\
SVHN      & $32\times32$ & 3 & 68,257 & 5,000 & 26,032 \\
CIFAR10  & $32\times32$ & 3 & 45,000 & 5,000 & 10,000 \\
\bottomrule
\end{tabular}
\vspace{-10px}
\end{table}

\subsection{Teacher Architectures}
\label{sect:methods:teacher}

Two binary teacher variants are evaluated because they expose two levels of binarization. Both use six binary convolutional blocks with channel widths $1\rightarrow128\rightarrow128\rightarrow256\rightarrow256\rightarrow512\rightarrow512$ for grayscale inputs and $3\rightarrow128\rightarrow\cdots\rightarrow512$ for color inputs, three max-pooling stages, four binary fully connected blocks $d_{\text{flat}}\rightarrow1024\rightarrow1024\rightarrow512\rightarrow256$, and a binary output layer $256\rightarrow10$. The penultimate fully connected layer produces the 256-dimensional pre-activation vector that is sign-binarized to form the LLAV.

The teachers differ in binarization strategy. The BCCNN binarizes weights through a straight-through estimator, while retaining ReLU activations between hidden layers. The BNN binarizes both weights and activations by replacing ReLU with a sign function at each hidden layer. Thus, the BCCNN preserves continuous activation magnitudes during intermediate forward computation, while the BNN operates on binary hidden signals after the first layer. Both teachers use BatchNorm after each binary layer. These variants separate the effect of weight-only binarization from the effect of fully binarized hidden computation.

\subsection{Training Protocol}
\label{sect:methods:protocol}

CRISP's protocol, as shown in Figure~\ref{fig:crisp_pipeline}, has two parts. The first establishes baseline CRISP reconstruction across all five datasets using the baseline 8-bit thermometer input encoding. The second isolates the effect of input Booleanization on reconstruction fidelity using SVHN and CIFAR10 configurations with binary thresholding, thermometer encoding, and quartile binning. For every dataset, the LLAV is extracted from the teacher's pre-logit layer and the raw input is Booleanized (Figure~\ref{fig:crisp_pipeline}c), forming the ITM targets and inputs, respectively. 
The ITMs (Figure~\ref{fig:crisp_pipeline}d) are trained on this Booleanized input for 5 epochs using fixed hyperparameters (50 clauses, threshold T = 20, specificity s = 4.0) instead of a per-dataset search, while data-dependent Booleanization thresholds are fitted on the training split only. ITMs run in parallel through a process pool, with progress checkpointed after each round.

\subsection{Evaluation Metrics}
\label{sect:methods:metrics}

The primary CRISP metric is reconstruction fidelity, defined in
Equation~\eqref{eq:reconstruction} as the mean per-neuron bit-level accuracy of
the ITM bank in reproducing the teacher LLAV, which is evaluated on the held-out test data.
The ITM mean accuracy is the average per-neuron training accuracy across the ITM bank, computed using Equation~\eqref{eq:itm_acc}:
{
    \vspace{-7.5px}
\begin{equation}
    \mathrm{Acc}_{\text{ITM}} = \frac{1}{H}\sum_{j=1}^{H} a_j,
    \label{eq:itm_acc}
    \vspace{-2.5px}
\end{equation}
}
where $a_j$ is the bit-level training accuracy of ITM $M_j$ and $H$ is the number of ITMs in the bank.
Note that this in-sample quantity is used only for selection, and is not reported as
a result. 
Teacher accuracy is reported to contextualize how informative the original teacher representation is, but the main CRISP objective is reconstruction of the teacher's internal state rather than direct replacement of the teacher classifier. The LLAV injection test additionally reports Cohen's $\kappa$ and the Matthews correlation coefficient (MCC) to assess agreement beyond raw accuracy in the reconstructed-head setting~\cite{dettori2020kappa,chicco2020advantages}. The Booleanization study reports the encoded input dimension and inference time to expose the cost of increasing the number of Boolean input bits.

\section{Results and Discussion}
\label{sect:results}

The results are organized around the three research questions. Section~\ref{sect:results:baseline} addresses RQ1 and RQ3 by reporting reconstruction behavior across datasets and teacher variants. Section~\ref{sect:results:llav_injection} isolates the effect of LLAV sign-binarization and ITM reconstruction error. Section~\ref{sect:Booleanization} addresses RQ2 by comparing Booleanization strategies on SVHN and CIFAR10. Section~\ref{sec:explainability} shows how the learned ITM clauses can be mapped back to input regions.

\subsection{Baseline Performances Across Datasets}
\label{sect:results:baseline}


Table~\ref{tab:baseline} reports CRISP's baseline performance across the five image classification benchmarks using two binary teacher architectures.
Both teachers share the same six-block convolutional structure and four-layer fully connected stack with a 256-bit LLAV, differing only in their binarization
strategy. 
The baseline reconstruction setting uses an 8-bit thermometer input encoding, and the reconstruction column reports the test fidelity, which follows Equation~\eqref{eq:reconstruction}: For each hidden neuron, the fraction of test samples on which the assigned ITM reproduces the teacher's binary activation, averaged over the 256 coordinates. This measurement is computed on held-out inputs rather than the samples used to train the ITMs, so it reflects generalisation rather than in-sample fit.
As a calibration check, the mean per-neuron majority-class baseline ranges from roughly 61\% to 74\% across datasets (mean 65\%), which the held-out fidelities exceed on most settings, so the reported values largely reflect learned reconstruction rather than trivial majority prediction. The exception is the low-accuracy BNN/SVHN teacher, whose reconstruction fidelity sits near its majority baseline.

\begin{table}[t]
    \centering
    \caption{CRISP performance across datasets and two binary teacher architectures using 8-bit thermometer input encoding.}
    \label{tab:baseline}
    \footnotesize
    \begin{tabular}{llcc}
        \toprule
        \textbf{Teacher}     & \textbf{Dataset}   & Teacher accuracy (\%) & ITM test fidelity (\%) \\
        \midrule
        BCCNN       & MNIST     & 99.65     & 85.99 \\
        BCCNN       & KMNIST    & 98.86     & 74.63 \\
        BCCNN       & FMNIST    & 94.20     & 83.92 \\
        BCCNN       & SVHN      & 95.01     & 67.77 \\
        BCCNN       & CIFAR10  & 91.06     & 71.54 \\
        \midrule
        BNN         & MNIST     & 98.98    & 84.31 \\
        BNN         & KMNIST    & 96.93    & 73.71 \\
        BNN         & FMNIST    & 87.83    & 85.10 \\
        BNN         & SVHN      & 73.10    & 73.12 \\
        BNN         & CIFAR10\textsuperscript{$\dagger$}  & 65.39    & 70.99 \\
        \bottomrule
    \end{tabular}
    \\ \vspace{3px}
    \begin{minipage}{\linewidth}
        \footnotesize \raggedright
        \textsuperscript{$\dagger$}The BNN/CIFAR10 teacher is a re-trained instance using the same BNN architecture described in Section~\ref{sect:methods:teacher}, as the original run submitted for peer review was not recoverable.
    \end{minipage}
    \vspace{-15px}
\end{table}

Two patterns follow from Table~\ref{tab:baseline}. 
First, on the datasets where the teachers are directly comparable, the teacher accuracy is higher for the BCCNN teacher, which indicates that retaining real-valued hidden activations between binary weight layers gives stronger task performance in the reported architecture. 
Second, symbolic reconstruction fidelity does not follow teacher accuracy monotonically, as the BNN produces comparable or higher held-out fidelity than the BCCNN on FMNIST and SVHN despite having lower or similar task accuracy. This distinction matters because CRISP measures recoverability of the internal code, and not only the quality of the teacher classifier.
In particular, a teacher can be accurate yet harder to reconstruct if its LLAV coordinates encode more distributed or spatially abstract evidence.

The MNIST-family datasets generally show higher reconstruction than the harder colour-image settings, as their class evidence is more closely tied to local stroke or shape patterns that can be expressed by pixel-threshold clauses, whereas SVHN and CIFAR10 require more spatial aggregation, which makes neuron-wise reconstruction from Booleanized pixels more difficult. 
KMNIST shows the lowest held-out fidelity relative to its teacher accuracy, and the injection test (Section~\ref{sect:results:llav_injection}) shows the same dataset losing the most classifier-head accuracy, which indicates that its hidden code generalises less readily to unseen inputs.
These results answer RQ1 by showing that the LLAV can be reconstructed on held-out data across all tested settings, with reduced fidelity as the visual task becomes less aligned with simple local evidence. They also answer RQ3 by showing that teacher architecture affects both task accuracy and reconstruction difficulty, but not always in the same direction.

\subsection{Effect of LLAV Binarization and Reconstruction Error}
\label{sect:results:llav_injection}

To isolate the cost of each stage in the symbolic pipeline, a three-way injection test was conducted using the BNN teacher and 8-bit thermometer input encoding. The teacher's final classifier head (\texttt{fc\_out}) was given three different inputs in turn which are the continuous pre-binarization activation vector of the penultimate layer, the sign-binarized LLAV, and the ITM reconstruction of that LLAV. Each input was passed through \texttt{fc\_out} alone, with predictions evaluated against ground-truth labels on the test set. 
This test is reported on the MNIST-family datasets, where the BNN teacher operates near its accuracy ceiling, and restricting to these datasets isolates the effect of reconstruction error on the classifier head from teacher error. 
Extending the injection test to the lower-teacher-accuracy SVHN and CIFAR10 settings is left for future work.

\begin{table}[]
    \centering
    \caption{Three-way LLAV injection test with the BNN teacher and 8-bit thermometer input encoding.}
    \label{tab:llav_injection}
    \footnotesize
    \begin{tabular}{llccc}
        \toprule
            \textbf{Dataset} & \textbf{Input to \texttt{fc\_out}} & \textbf{Accuracy (\%)} & $\kappa$ & \textbf{MCC} \\
        \midrule
            \multirow{3}{*}{MNIST}
             & BNN baseline       & 98.98 & 0.9887 & 0.9887 \\
             & LLAV binarized     & 98.98 & 0.9887 & 0.9887 \\
             & ITM reconstruction & 84.68 & 0.8297 & 0.8302 \\
        \midrule
            \multirow{3}{*}{FMNIST}
             & BNN baseline       & 87.83 & 0.8648 & 0.8649 \\
             & LLAV binarized     & 87.83 & 0.8648 & 0.8649 \\
             & ITM reconstruction & 78.41 & 0.7601 & 0.7623 \\
        \midrule
            \multirow{3}{*}{KMNIST}
             & BNN baseline       & 96.93 & 0.9659 & 0.9659 \\
             & LLAV binarized     & 96.93 & 0.9659 & 0.9659 \\
             & ITM reconstruction & 58.80 & 0.5422 & 0.5422 \\
        \bottomrule
    \end{tabular}
    \vspace{-10px}
\end{table}

The injection test reveals two findings. First, sign-binarizing the LLAV incurs no accuracy cost in the reported rows, as the continuous and binary conditions are identical. This follows from the structure of \texttt{fc\_out} as a \texttt{BinLinear} layer, it applies a sign operation before the linear transform and therefore discards magnitude information at this interface. In this setting, the binary LLAV is sufficient for reproducing the teacher head. Second, ITM reconstruction reduces accuracy across all reported datasets, with the largest drop on KMNIST. The main accuracy loss comes from imperfect reconstruction of the sign pattern, not from LLAV sign-binarization itself.
 
A reconstruction error in one LLAV coordinate changes the input seen by \texttt{fc\_out}. Since the output layer uses binarized weights and a sign-encoded input, flipping one LLAV bit changes each class logit by a fixed signed amount determined by the corresponding binary output weight~\cite{rastegari2016xnor}. Accumulated bit errors across the 256-dimensional LLAV can therefore change the class margin and produce a misclassification. This reframes CRISP's bottleneck as the fidelity of the ITM bank in reproducing the teacher LLAV. 

\begin{figure*}[]
    \centering
    \begin{subfigure}[t]{0.45\textwidth}
        \centering
        \includegraphics[width=\linewidth]{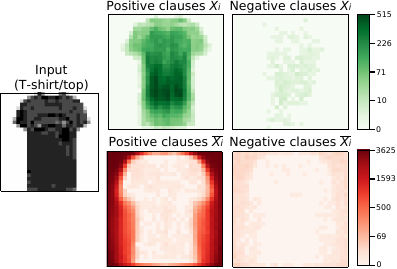}
        \caption{FMNIST T-shirt/top.}
        \label{fig:fmnist-collective}
    \end{subfigure}
    \hspace{10px}
    \begin{subfigure}[t]{0.45\textwidth}
        \centering
        \includegraphics[width=\linewidth]{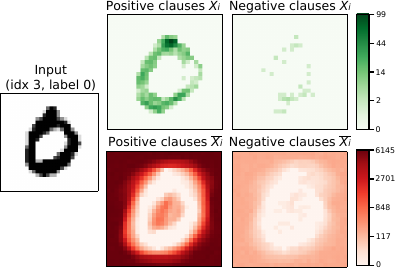}
        \caption{MNIST digit ``0''.}
        \label{fig:mnist-collective}
    \end{subfigure}
    \caption{Collective clause evidence pooled across all 256 ITMs for representative test samples. Each panel separates positive- and negative-polarity clauses and maps positive literals $X_i$ and negated literals $\overline{X_i}$ back to image coordinates. Color intensity denotes pooled literal count within each row.}
    \label{fig:collective}
    \vspace{-10px}
\end{figure*}

\subsection{Effect of Input Booleanization}
\label{sect:Booleanization}

\begin{table}[]
    \centering
    \caption{Effect of Booleanization strategy on ITM reconstruction fidelity
    for SVHN and CIFAR10.}
    \label{tab:Booleanization}
    \footnotesize
    \begin{tabular}{ccccccc}
        \toprule
        \makecell{\textbf{Dataset}} & \makecell{\textbf{Enc.}} & \makecell{\textbf{Bits}} & \makecell{\textbf{Input}\\\textbf{Bits}} & \makecell{\textbf{Train}\\\textbf{acc. (\%)}} & \makecell{\textbf{Test}\\\textbf{fid. (\%)}} & \makecell{\textbf{Inf.}\\\textbf{Time (s)}} \\
        
        \midrule
        SVHN & Bin & 1 & 3072  & 82.94 & 86.03 & 275.8 \\
        SVHN & Th  & 1 & 3072  & 82.66 & 85.70 & 274.4 \\
        SVHN & Th  & 2 & 6144  & 79.19 & 79.98 & 499.1 \\
        SVHN & Th  & 4 & 12288 & 81.49 & 83.82 & 1231.0 \\
        SVHN & Q   & 1 & 3072  & 84.93 & \textbf{87.52} & 275.7 \\
        SVHN & Q   & 2 & 6144  & 82.16 & 84.06 & 414.1 \\
        SVHN & Q   & 4 & 12288 & 82.05 & 83.02 & 1056.0 \\
        \midrule
        CIFAR10 & Bin & 1 & 3072  & 92.31 & 92.08 & 148.6 \\
        CIFAR10 & Th  & 1 & 3072  & 92.29 & 92.04 & 129.3 \\
        CIFAR10 & Th  & 2 & 6144  & 91.32 & 91.45 & 214.3 \\
        CIFAR10 & Th  & 4 & 12288 & 91.37 & 91.47 & 542.0 \\
        CIFAR10 & Q   & 1 & 3072  & 92.47 & \textbf{92.12} & 84.1 \\
        CIFAR10 & Q   & 2 & 6144  & 91.28 & 91.39 & 167.0 \\
        CIFAR10 & Q   & 4 & 12288 & 91.44 & 91.54 & 464.5 \\
        \bottomrule
    \end{tabular}
    \vspace{-10px}
\end{table}

Table~\ref{tab:Booleanization} reports the effect of Booleanization strategy on SVHN and CIFAR10. The table includes the encoding (Enc.), bit depth, encoded input dimension (Input Bits), mean ITM training accuracy (Train acc.), test-set reconstruction fidelity (Test fid.), and inference time (Inf. time). Binary thresholding (Bin) is evaluated only at one bit because it produces one Boolean feature per scalar input, while thermometer (Th) and quartile (Q) encodings are evaluated at one, two, and four bits.

Two patterns are visible in Table~\ref{tab:Booleanization}. 
First, reconstruction fidelity at one-bit depth is close across all three schemes. On SVHN, test fidelity ranges from 85.70\% to 87.52\%, while on CIFAR10, it ranges from 92.04\% to 92.12\%. This similarity is expected, as all three one-bit schemes reduce each scalar feature to a single threshold test. The threshold is fixed at 0.5 for binary and thermometer encodings, and fitted as a training-set median for quartile binning. 
In particular, the one-bit binary thresholding (Eq.~\ref{eq:binary}) and one-bit thermometer encoding (Eq.~\ref{eq:therm}) are identical by construction, since both reduce to $\mathbf{1}[u>0.5]$. The small difference between their rows in Table~\ref{tab:Booleanization} (e.g. 86.03\% vs.\ 85.70\% on SVHN) reflects only run-to-run stochasticity in ITM training, which indicates that sub-point differences elsewhere in the table are within noise.
Second, quartile binning gives the strongest one-bit result on SVHN (87.52\%) where the margin exceeds run-to-run variation. For CIFAR10, the one-bit schemes are effectively tied (92.12\% vs.\ 92.08\%) meaning no encoding is singled out.

Fidelity does not improve monotonically with bit depth. For both thermometer and quartile encodings, most two-bit and four-bit settings are lower than the corresponding one-bit setting. 
As noted after Equation~(5), the highest thermometer bit is constant by construction, so an $n_{\text{bits}}$ code carries fewer informative bits than its dimension suggests, and the input dimension grows faster than the useful threshold information available to the ITMs. 
The added dimensions also enlarge the clause search space, which can reduce reconstruction quality under a fixed training budget.


The timing results show the cost of wider encodings. On SVHN, thermometer inference time rises from 274.44s at one bit to 1231.02s at four bits, while quartile inference time rises from 275.65s to 1056.01s over the same range. CIFAR10 follows the same trend, but the absolute timings are lower. 
These are total wall-clock times over the full test set (Table I), where per-image latency is on the order of 10–50ms across the encodings.
These results answer RQ2 by showing that richer Booleanization does not automatically improve LLAV reconstruction. In the reported experiments, one-bit quartile binning provides the best fidelity-cost trade-off.

\subsection{Explainable Symbolic Evidence from ITMs}
\label{sec:explainability}

The preceding results quantify whether the ITM bank reconstructs the teacher's LLAV. 
This section applies clause-evidence pooling, aggregating across all 256 ITMs, to representative test samples.
The two examples in Figure~\ref{fig:collective} use 8-bit thermometer encoding on MNIST and FMNIST, matching the baseline reconstruction setting in Section~\ref{sect:results:baseline}.

Although each ITM is trained on a flat vector of pixel-threshold features, the pooled evidence in Figure~\ref{fig:collective} forms spatially coherent maps. This structure arises from training pressure rather than an explicit spatial prior. An ITM retains literals that help predict its assigned LLAV bit and discards literals that do not. In centered image datasets, pixels on or near the object are more predictive than uniform background pixels, so pooled literals concentrate around the object region.

The MNIST digit in Figure~\ref{fig:mnist-collective} provides a useful check because a ``0'' contains two background regions, namely, the outside of the digit and the hole inside the loop. The pooled $\overline{X_i}$ evidence fills the interior hole as well as the exterior background, which is consistent with local pixel-threshold evidence rather than a simple boundary-only pattern. The FMNIST example in Figure~\ref{fig:fmnist-collective} similarly concentrates positive-literal evidence over the T-shirt silhouette.

The raw scale of the $X_i$ and $\overline{X_i}$ maps should not be compared directly. The negated-literal maps have larger counts because thermometer encoding allows near-zero background pixels to satisfy several low-threshold negations, while foreground pixels activate a narrower threshold band near their intensity. Thus, the spatial location of evidence is interpretable, but the magnitude gap between positive and negated literal maps should not be treated as a direct measure of foreground versus background importance. Negative-polarity clauses contribute less evidence in the two displayed samples, which is consistent with confident correct predictions where supporting clauses for the winning class carry most of the reconstruction signal.

\subsection{Limitations}

All results use a single training run per configuration, where training the full 256-ITM bank is memory-bound as a complete run on CIFAR10 over all training samples exceeds 24 hours per configuration on our setup, which precludes multi-seed repetition and formal uncertainty estimation. 
Small inter-encoding differences should therefore be read with caution, as the 0.04-point quartile-over-binary advantage on CIFAR10 detailed in Table~\ref{tab:Booleanization} lies within run-to-run variation and is not claimed as meaningful, whereas the larger SVHN gap is more robust.
The per-coordinate ITMs are individually small, so the dominant cost is repeated evaluation of all training samples across the bank rather than ITM size; sample subsampling, minibatched clause updates, and scaling the bank beyond 256 coordinates are left to future work. Reconstruction fidelity in Table~\ref{tab:baseline} is reported on held-out test data following Eq.~\eqref{eq:reconstruction}.

\section{Conclusion}
\label{sect:conclusion}

This paper presented CRISP, a neuro-symbolic framework that reconstructs the last-layer activation vector of a binary neural network as propositional clauses, assigning one ITM to each hidden neuron over a Boolean encoding of the input (Section~\ref{sect:crisp}). The surrogate preserves the identity of every reconstructed coordinate and exposes its supporting literals, turning the teacher's final hidden interface into an inspectable symbolic object. Experiments across five image benchmarks and two binary teacher architectures show that reconstruction fidelity is governed by task difficulty and teacher binarization (Sections~\ref{sect:results:baseline} and~\ref{sect:results:llav_injection}). Binarization alone does not reduce teacher-head accuracy in the tested BNN setting, so the dominant loss stems from imperfect ITM reconstruction, while the Booleanization study (Section~\ref{sect:Booleanization}) shows that one-bit quartile binning, rather than higher bit depth, yields the strongest fidelity-cost trade-off, most clearly on SVHN.

Within the reported scope, LLAV reconstruction is feasible but harder on complex colour datasets (RQ1), one-bit quartile binning gives the best fidelity-cost balance (RQ2), and teacher architecture affects task accuracy and reconstructability independently (RQ3). Clause pooling further shows reconstructed bits forming spatial evidence maps despite the ITMs having no explicit spatial prior.


\section{Future Work}

Several directions follow from this work. 
First, reconstruction could extend from the final hidden interface to intermediate layers, toward a layer-by-layer clause-level account of the whole network, potentially with reconstruction-aware teacher training so hidden representations become recoverable by construction. 
Second, clause banks could be coupled with satisfiability-based verification to state symbolic guarantees for safety-critical deployment. 
Finally, the approach could scale to larger color benchmarks with channel-specific RGB evidence and imbalance-aware baselines, and generalize to tabular, sensor, biomedical, and time-series data, where a logic-native clause bank offers an accurate, low-complexity, and explainable alternative.

\section*{Acknowledgment}
This work was partially supported by the EPSRC EP/X036006/1 Scalability Oriented Novel Network of Event Triggered Systems (SONNETS) project.

\bibliographystyle{IEEEtran}
\bibliography{references}


\end{document}